\documentclass[10pt,twocolumn,letterpaper]
{article}

\usepackage{cvpr}              
\usepackage{booktabs}
\usepackage{multirow}
\usepackage{graphicx}
\usepackage{caption}
\usepackage{xcolor}
\usepackage{colortbl}
\definecolor{mygray}{gray}{0.92}
\usepackage{amsfonts}
\usepackage{todonotes}
\usepackage{gensymb}
\usepackage{amssymb}
\usepackage{bbding}
\usepackage{textcomp}
\usepackage{amsmath}
\usepackage{array}
\usepackage{pifont} 
\newcommand{\cmark}{\ding{51}} 
\newcommand{\xmark}{\ding{55}}  
\usepackage{xcolor}
\definecolor{baselinecolor}{gray}{0.8}
\usepackage{tabularx}
\usepackage{multicol}
\usepackage{listings}
\definecolor{lightgray}{gray}{0.9}

\makeatletter

\newcommand{\Rmnum}[1]{\expandafter\@slowromancap\romannumeral #1@}
\makeatother

\newcolumntype{x}[1]{>{\centering\arraybackslash}p{#1pt}}
\newcolumntype{y}[1]{>{\raggedright\arraybackslash}p{#1pt}}
\newcolumntype{z}[1]{>{\raggedleft\arraybackslash}p{#1pt}}

\usepackage{subcaption}
\usepackage{wrapfig}
\usepackage[accsupp]{axessibility}

\definecolor{cvprblue}{rgb}{0.21,0.49,0.74}
\usepackage[pagebackref,breaklinks,colorlinks,allcolors=cvprblue]{hyperref}

\def\paperID{14832} 
\def\confName{CVPR}
\def\confYear{2025}

\title{Period-LLM: Extending the Periodic Capability of \\ Multimodal Large Language Model}

\author{Yuting Zhang$^{1\ast}$, Hao Lu$^{1,2\ast}$, Qingyong Hu$^{2}$, Yin Wang$^{3}$, Kaishen Yuan$^{1}$, Xin Liu$^{4}$, Kaishun Wu$^{1\ddagger}$\\
$^{1}$The Hong Kong University of Science \& Technology (Guangzhou), \\
$^{2}$The Hong Kong University of Science \& Technology,
$^{3}$Zhejiang University, \\
$^{4}$Lappeenranta-Lahti University of Technology\\
\{yzhang430, hlu585\}@connect.hkust-gz.edu.cn, wuks@hkust-gz.edu.cn\\
$^\ast$ Equal contribution, $^\ddagger$ Corresponding author
}

\begin{document}
\twocolumn[{
	\renewcommand\twocolumn[1][]{#1}
	\maketitle
        \vspace{-22mm}
	\begin{center}
		\centering
		\includegraphics[width=1\linewidth]{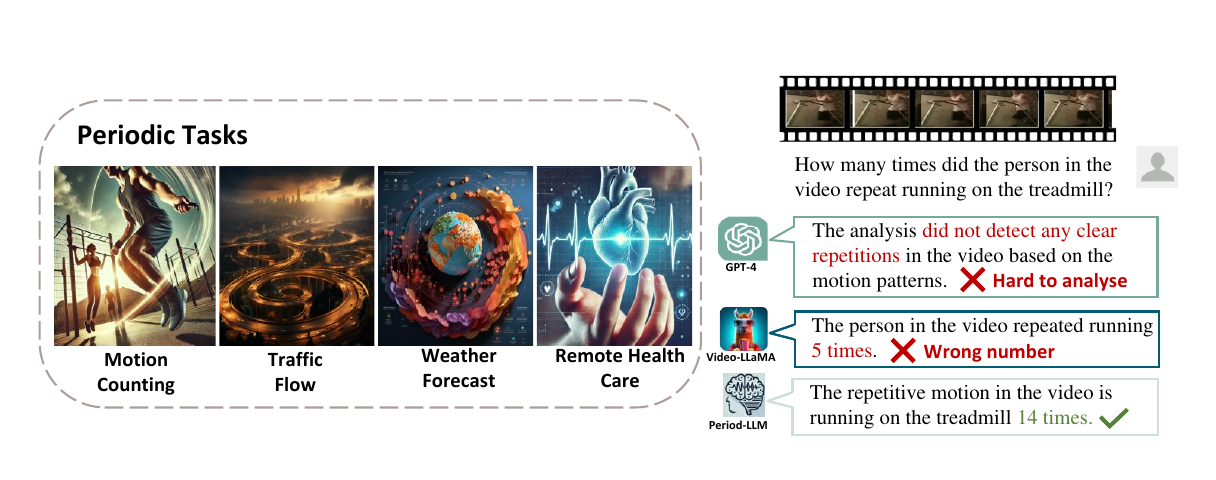}
			\vspace{-3.5em}
		\captionof{figure}{Existing multimodal models may fail to analyze periodic tasks, such as motion counting, traffic flow, weather forecasting, and remote health care. For example, GPT-4 fails to detect clear repetitions due to the difficulty in analyzing motion patterns, and Video-LLaMA incorrectly counts the repetitions. In contrast, Period-LLM accurately analyzes the actions and provides correct repetition count.}
		\label{fig:performance1}
	\end{center}
}]
\begin{abstract}
\vspace{-2.5em}


Periodic or quasi-periodic phenomena reveal intrinsic characteristics in various natural processes, such as weather patterns, movement behaviors, traffic flows, and biological signals. Given that these phenomena span multiple modalities, the capabilities of Multimodal Large Language Models (MLLMs) offer promising potential to effectively capture and understand their complex nature. However, current MLLMs struggle with periodic tasks due to limitations in: 1) lack of temporal modelling and 2) conflict between short and long periods. This paper introduces Period-LLM, a multimodal large language model designed to enhance the performance of periodic tasks across various modalities, and constructs a benchmark of various difficulty for evaluating the cross-modal periodic capabilities of large models. Specially, We adopt an ``Easy to Hard Generalization" paradigm, starting with relatively simple text-based tasks and progressing to more complex visual and multimodal tasks, ensuring that the model gradually builds robust periodic reasoning capabilities. Additionally, we propose a ``Resisting Logical Oblivion" optimization strategy to maintain periodic reasoning abilities during semantic alignment. Extensive experiments demonstrate the superiority of the proposed Period-LLM over existing MLLMs in periodic tasks. The code is available at \url{https://github.com/keke-nice/Period-LLM}.

\end{abstract}    
\vspace{-2.5em}
\section{Introduction}
\label{sec:intro}

Periodic tasks refer to activities performed at fixed or predictable intervals, encompassing a wide range of applications. At a macro level, this includes repetitive motions in human daily life, such as step counts and the number of times one jumps rope~\cite{dwibedi2020counting, zhang2021repetitive, hu2022transrac}, as well as the periodic analysis of meteorological data, where scientists can identify various weather patterns and trends, forecast future weather conditions~\cite{rodrigues2018deepdownscale, abhishek2012weather, scher2018predicting}, and issue timely warnings.
At a micro level, in health monitoring, by capturing and analyzing subtle chromatic changes in facial videos, physiological signals such as heart rate, blood pressure, and respiratory rate can be extracted~\cite{liu2024rppg, lu2021dual, yang2022simper, zhang2024advancing} . The periodic fluctuations of these signals reflect the physiological state of human. Effectively handling periodic tasks can significantly enhance various aspects of human life, from predicting weather patterns to monitoring health conditions.

Given that periodic phenomena occur across various modalities, it becomes essential to utilize multimodal models for period analysis instead of single modality. In recent years, multimodal large language models~\cite{videollama, liu2024visual_llava, llamavid, alayrac2022flamingo} have advanced the progress of visual and linguistic learning by incorporating visual encoders into various pre-trained large language models. Furthermore, some studies have extended vision to other modalities such as audio~\cite{videollama, han2024onellm}, depth imagery~\cite{girdhar2023imagebind}, and near-infrared~\cite{girdhar2023imagebind}. With the promising progress of LLM-enabled multimodal models, a natural question arises: \textit{Can multimodal large language models understand periodic tasks?}

Unfortunately, existing large models have shown inadequacies in counting capabilities or understanding ``repetitive" temporal sequences~\cite{li2024mvbench}, creating obstacles for the application of periodic tasks in multimodal large language models, as shown in Fig.~\ref{fig:performance1}. These shortcomings mainly stem from the following challenges:

\textbf{1) Interference of spatial pseudo-temporal information.} The leakage of numerical information in visuals may lead models to take shortcuts rather than learning the true periodic information. When specific numbers appear in videos, models may tend to directly answer with these numbers, or be misled by them, rather than deriving answers through the analysis of periodic information~\cite{paiss2023teaching}. This limits the models' generalization capabilities, rendering them unable to handle periodic tasks without direct numerical information.
\textbf{2) Conflict between long-time periodic reasoning ability and short-time semantic alignment. } The recognition of a periodic task lies not only in the ability to deduce the number of periodic repeats, but also in the accurate recognition of the semantic information in the video. Both require the model to have the ability to model the timing, but with different focus. Too much focus on the short-time semantic understanding may lead to ignorance of periodic reasoning ability.
\textbf{3) Absence of captions} is a significant issue. In the training corpora of MLLMs, most sentences are descriptive and rarely contain specific descriptions of the number of actions. This poses difficulties for models when dealing with periodic tasks that require counting the number of actions. For example, a model might respond with ``The boy in the video is doing pull-ups" or ``The boy in the video does multiple pull-ups", without a specific counting supervision.

To avoid overfitting to modality-specific features while preserving temporal modeling capabilities, we propose a MLLM with periodic capabilities, termed Period-LLM. Specifically, we employ an easy-to-hard generalization paradigm. \textbf{Initially}, we define text-based, macro, and micro periodic tasks, ranging from easy to hard. We anticipate starting with text and progressively advancing, enabling Large Language Models (LLMs) to learn periodicity across different modalities. The rationale for this definition is twofold: 1) Despite modality alignment, LLMs still exhibit superior text processing capabilities compared to other modalities; 2) Macro periodic tasks contain clear semantic features, making their periodic nature easier to capture than that of micro tasks. Accordingly, we selected periodic tasks corresponding to different levels of difficulty. At the textual level, we construct a ``repeated word question-answer" dataset to pre-train LLMs, deepening their understanding of \textit{``repetitiveness"} from a textual perspective. At the macro level, we select the repetitive motion dataset Countix~\cite{dwibedi2020counting}, which aims to predict the number of repetitive movements in videos. At the micro level, we choose the rPPG task, which involves capturing subtle periodic or quasi-periodic physiological signals from facial videos or radio-frequency signals. 

\textbf{Furthermore}, it is believed that a good periodic model should have both the ability of periodic reasoning and the ability of semantic understanding. Focusing on the optimization of semantic alignment may, to some extent, lead to forgetting the logical reasoning abilities learned from simple tasks. Therefore, we propose a \textit{Resisting Logical Oblivion} optimization strategy. Specifically, by manually designing a method to dynamically update the backpropagation gradients, the model can adjust previously overlooked feature dimensions without interfering with existing ones, providing a guided approach to model optimization. A series of experiments demonstrate the superiority of the proposed Period-LLM over existing MLLMs in periodic tasks. This advantage is not limited to text and video; we further explore the potential of other modalities based on this foundation. 

Our contributions are summarized as follows:
 
\begin{itemize}
\item To the best of our knowledge, we propose the first MLLM-based periodic model (Period-LLM) and construct a benchmark featuring periodic tasks of varying difficulty, including text, video, and other modalities.
\item We propose an easy-to-hard training paradigm that progresses from simple to complex tasks, enabling the model to sequentially acquire the ability to perceive periodic information across different modalities, starting with text, then vision, and extending to other modalities.
\item We propose a Resisting Logical Oblivion (RLO) optimization strategy to resolve the conflict between periodic reasoning and semantic alignment.
\item Extensive experiments demonstrate that the proposed Period-LLM outperforms existing MLLMs in periodic tasks across various modalities.

\end{itemize}

\begin{figure*}
\centering
\includegraphics[width=\linewidth]{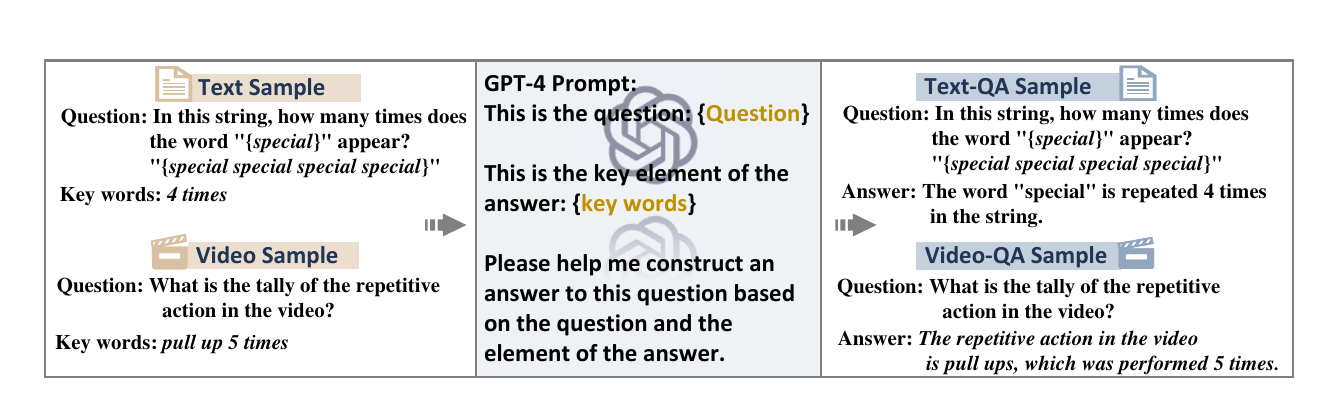}
\vspace{-3em}
  \caption{\small{The process of the generation of Question-and-answer datasets. Samples of text and video are taken as examples. In the text sample, questions and the occurrences of specific words are randomly generated, while in the video sample, the questions are randomly generated and the keywords represent labels within the dataset. These elements are then incorporated into a GPT-4 prompt to construct answers that include key information. The right side of the image shows examples of the final Q\&A pairs generated, where the answers are complete sentences containing the critical information from the key elements.
  }
  }
\vspace{-5mm}
\label{fig:QA_generation}
\end{figure*} 
\vspace{-1em}
\section{Related Work}
\label{related work}
\subsection{Multimodal Large Language Model}
In recent studies, large models have exhibited reliable capabilities in the field of video understanding. In addition to the powerful vision-language large models~\cite{li2023blip,zhu2023minigpt,liu2024visual,lu2024gpt}, there is an increasing focus on exploring more modalities in recent research~\cite{lv2024video,li2023videochat,maaz2023video,ye2023mplug,luo2023valley}.
Bain et al.~\cite{Bain21} introduce a large-scale dataset, which provides general descriptions associated with video content. Several LLM-based works~\cite{li2023videochat,maaz2023video,ye2023mplug,luo2023valley} are proposed to comprehend the visual content of the video. Besides, Video-LLaMa~\cite{damonlpsg2023videollama} extends the ability to comprehend both auditory and visual information, while Su et al.~\cite{su2023pandagpt} leverages the power of multimodal encoders to understand across six modalities. Recently, Lv et al.~\cite{lv2024video} proposed video-based large language models for the task of VAD in a weakly supervised framework. Different from them, our work focuses on the periodic analysis of MLLMs.
\subsection{Periodic Tasks}
Examples of periodic learning include recovering and magnifying physiological signals (e.g., heart rate or breathing) \citep{niu2019rhythmnet,lu2021dual}, predicting weather and environmental changes (e.g., nowcasting of precipitation or land surface temperatures) \citep{sonderby2020metnet}, counting motions that are repetitious (e.g., exercises or therapies) \citep{dwibedi2020counting,ali2020spatio}, and analyzing human behavior (e.g., gait) \citep{liu2022monitoring}. Prior work has focused on designing customized neural architectures \citep{liu2020multi,dwibedi2020counting}, loss functions \citep{starke2022deepphase}, and leveraging relevant learning paradigms including transfer learning \citep{lu2018class} and meta-learning \citep{liu2021metaphys} for periodic learning. This paper attempts to study the capability of periodic tasks in multimodal models.

\vspace{-0.6em}
\section{Problem Setting and Preliminaries}
\label{sec:problem}

\begin{figure*}
\centering
\includegraphics[width=0.9\linewidth]{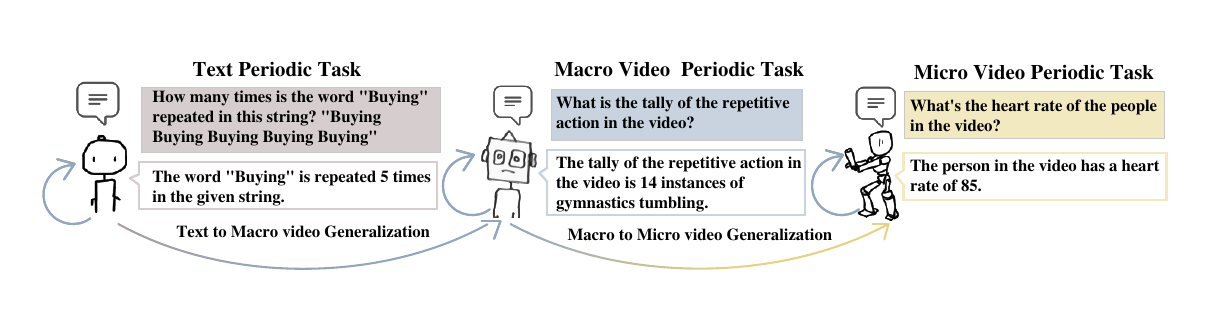}
\vspace{-1.8em}
  \caption{\small{The pipeline of training paradigm from easy to hard. The model begins training with simpler periodic tasks and gradually transitions to more complex ones. Throughout this process, Period-LLM's understanding of periodic tasks progressively strengthens and becomes more profound.
  }
  }
 \vspace{-1.6em}
\label{fig:pipeline}
\end{figure*}

In nature, periodic phenomena contain both periodic and semantic information, which can be modeled by \( x = K \cdot p(\omega t) + N \cdot \text{s}(t) \), where \( p(\omega t) \) is a periodic function with angular frequency \( \omega = 2\pi/T \) and period \( T \). Here, \( K \) represents the amplitude of the periodic component, \( N \) denotes the amplitude of the semantic component, and \( \text{s}(t) \) is a semantic function related to time \( t \).
When discretizing continuous sequences as model inputs \( X \), we adopt a sampling interval \( \Delta t = T/n \), where \( n \) is a positive integer multiple of the sampling rate (\( n = 1, 2, 3, \ldots \)). Ideally, when \( n = 1 \) and semantics are absent, the value at each sampling point equals the constant \( K \), resulting in \( x = K \). This can be mathematically expressed as:
\vspace{-0.4em}
\begin{equation}
    X = 
\begin{cases}
    K \cdot p(\omega t) + N \cdot \text{s}(t), & \Delta t = \frac{T}{n}, \; n = 2, 3, 4, \ldots \\
    K, & \Delta t = T 
\end{cases}
\end{equation}
However, in practical applications, data is often high-dimensional and noisy. For instance, after processing a text segment through a tokenizer, we obtain a matrix with dimensions \([L, \text{Dim}]\), where \( L \) represents the length of the text (i.e., the number of tokens), and \( \text{Dim} \) denotes the embedding dimension of each token. In cases of repetitive text such as “\textit{period period period}”, certain dimensions of the encoded feature matrix may exhibit constant values similar to \( x = K \), where \( K \) is a vector representing the embedding values of these tokens in specific dimensions.

Inputs from other modalities, such as video, present more complex scenarios. Periodic movements may be obscured by a large amount of semantic or redundant information. In macro-periodic tasks, periodic movement is more prominent, allowing models to detect periodicity more easily; however, in micro-periodic tasks, the amplitude of the periodic signal (\( K \)) may be small and masked by noise, making it challenging for models to capture these subtle periodic variations.
Despite these complexities, research suggests that models can transfer knowledge from simpler tasks to more complex ones. In this context, repetitive text can be seen as a basic periodic task, while macro-periodic and micro-periodic video tasks represent more complex periodic challenges. Although “repetition” is the common thread across these tasks, the semantic content and amplitude of the periodic signal vary.

Given that large language models (LLMs) have been trained on extensive and diverse text corpora, they are particularly well-suited for processing text tasks. We propose the following framework:
\vspace{-0.8em}
\begin{equation}
    A = F(X, Q)
\end{equation}
\vspace{-1em}

This framework accommodates inputs \( X \) from various modalities, where \( X \) represents periodic data from any modality, \( Q \) represents a given question, and the model computes the answer \( A \). We suggest beginning model training with simple repetitive text tasks and gradually progressing to macro-periodic and micro-periodic video tasks. This progressive training approach helps the model develop a comprehensive understanding of periodic phenomena, moving from simple to complex tasks, thereby enhancing its ability to handle periodic challenges effectively.

\section{Method}
In this paper, we aim to enhance the understanding of large models regarding periodic tasks. Given the current absence of specialized question-and-answer datasets for periodic tasks and corresponding testing standards, we will undertake pioneering work in the content that follows. Specifically, we firstly elucidate the instruction generation for periodic tasks. Then, we explore the generalization process from simpler tasks to more complex ones. Additionally, we propose an optimization strategy to address the challenge of preserving periodic reasoning abilities during the semantic alignment phase of Multi-Modal Generalization.

\vspace{-0.3em}
\subsection{Instruction Generation for Periodic Tasks}
\label{sec:Instruction}

\subsubsection{Text-Only Instructions} 

To enhance our dataset's diversity, we use GPT-4 to help create a repetitive text dataset through several key steps:
First, we define the core question: ``How many times is the word \{\textit{my word}\} repeated in the string \{\textit{my string}\}?" We then use GPT-4 to generate ten semantically similar questions, \( q_i \in \{q_1, q_2, \ldots, q_{10}\} \).
For word diversity, we randomly select a word from the GPT-4 technical report~\cite{achiam2023gpt} as \{\textit{my word}\}. To further diversify, we set a repetition count \( n \) within \{2, 3, ..., 20\}, constructing \{\textit{my string}\} as ``\(\{\textit{selected word}\} \times n\)".
Since GPT-4 can't directly calculate word repetitions, we guide it with precise prompts. We instruct GPT-4: ``Construct an answer based on the question format, selected word, and repetition count." During this process, we randomly choose a question \( q_i \) and provide the necessary answer elements, \{selected word, \( n \) times\}, enabling GPT-4 to generate a fitting response. 
By following these steps, we effectively use GPT-4's generative capabilities to build a diverse and accurate repetitive text dataset, which supports robust model training and evaluation.

\subsubsection{Multi-Modal Instructions} 

For tasks involving periodic data across various modalities like video, traffic streams, and health monitoring, we employ a method that integrates annotations, raw captions, and frequency information to generate comprehensive representations. This approach is applicable to diverse datasets, addressing challenges such as dynamic environments, variations in periodicity and counts, and action speeds.
We transform data into a question-and-answer format inspired by repetitive text dataset construction. We generate unified captions for each segment by combining existing annotations and frequency data. The main question posed is: ``What is the total number of repetitive actions in this data?"
Using GPT-4's capabilities, we craft multiple semantically similar questions from this core inquiry. Explicit instructions and answer elements are provided to generate question-and-answer pairs, enhancing dataset representation, the process as shown in \cref{fig:QA_generation}. This method enriches model understanding and processing of periodic tasks across modalities.
We establish a link between repetitive text and periodic data, enabling generalization from simple to complex tasks. The process first involves self-optimization in relatively simple individual tasks, followed by fine-tuning on more complex tasks, as illustrated in \cref{fig:pipeline}.

\begin{figure*}
\centering
\includegraphics[width=0.9\linewidth]{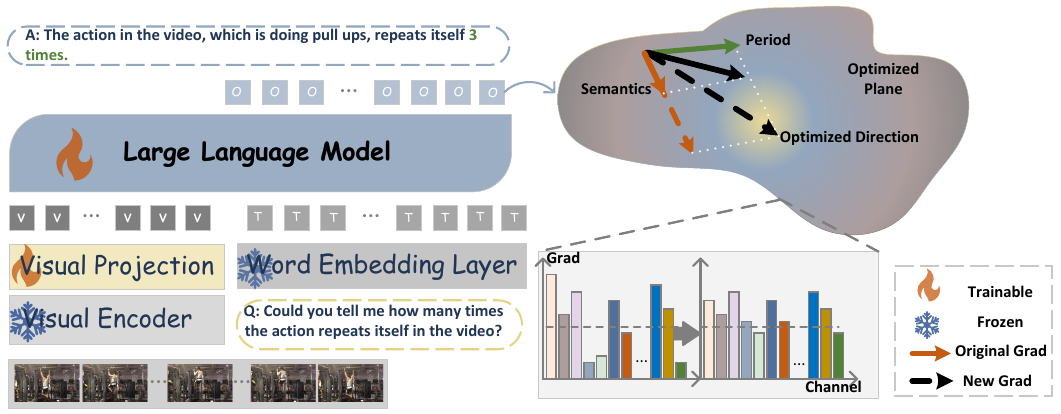}
\vspace{-1em}
  \caption{\small{The Period-LLM framework, using video as an example, integrates both text and video inputs. The video is processed through a visual encoder and a visual projector, then concatenated with text features derived from a word embedding layer. This combined input is fed into a large language model (LLM). Gradient optimization strategies are applied to the output feature channels of the LLM, specifically targeting those channels that failed to capture useful information, thereby enriching the semantic content in these channels.}}
\vspace{-1.5em}
\label{fig:framework}
\end{figure*} 

\subsection{Easy-to-Hard Generalization}
\label{sec:Generalization}

In this section, we explore the generalization process from simpler tasks to more complex ones, structured around the problem definition \( A = F(f, Q) \), where \( A \) represents the answer, \( f \) represents the features derived from the input data, and \( Q \) represents the question.

\subsubsection{Text-Only Generalization}

The first stage of generalization focuses on text-based tasks, where the goal is to learn logical reasoning without needing to align other modalities with text. This can be mathematically represented as:
\vspace{-1em}
\begin{equation}
    A = F(T_f, Q),
\end{equation}
here, \( T_f \) represents the features derived from textual input. The emphasis is on developing the model's logical reasoning capabilities, as the input modality is solely text, which simplifies the task by eliminating the need to handle multi-modal alignment. The model learns to process and understand the logical structure of questions and generate appropriate answers purely based on textual information.

\subsubsection{Multi-Modal Generalization}

The next stage involves more complex tasks that require the integration of multiple modalities, such as video and text. This stage not only demands logical reasoning but also the alignment of semantic information across different modalities. This process can be described by:
\vspace{-0.8em}
\begin{equation}
    A = F(M_f, Q),
\end{equation}
in this case, \( M_f \) represents features derived from a combination of inputs, including video and text. The challenge is twofold: the model must align semantic content across different modalities while simultaneously developing robust periodic logical reasoning capabilities. This complex task requires the model to integrate semantic features with textual information and establish accurate connections between them in order to answer questions correctly.

\subsection{Resisting Logical Oblivion}
\label{sec:self}


Due to the significant differences between periodic logical reasoning abilities and the ability to understand semantic information in videos, we design a Resisting Logical Oblivion (RLO) optimization strategy to prevent the new knowledge learned in a new round of optimization from overwriting the reasoning knowledge acquired from simpler tasks. Vanilla gradient descent leads to an undifferentiated update of the entire feature space. If \(\theta = (\theta_1, \theta_2, \ldots, \theta_J) \in \mathbb{R}^{C \times W}\) is the parameter of the model, where \(W\) is related to the network structure, \(\frac{\partial L(\theta)}{\partial \theta_j}\) (\(j = 1, 2, \ldots, J\)) represents the partial derivative of \(\theta\). The process of model updating is as follows:
\vspace{-1em}
\begin{align}
\label{eq:5}
\nabla \theta_j &= -\alpha \frac{\partial L(\theta)}{\partial \theta_j}, \\
\theta'_j &\leftarrow \theta_j + \nabla \theta_j,
\end{align}
where \(\alpha\) is the learning rate, \(\nabla \theta_j\) is the updating vector, and \(\theta'_j\) is the new weight for the next iteration.

\begin{figure}
\centering
\includegraphics[width=0.95\linewidth]{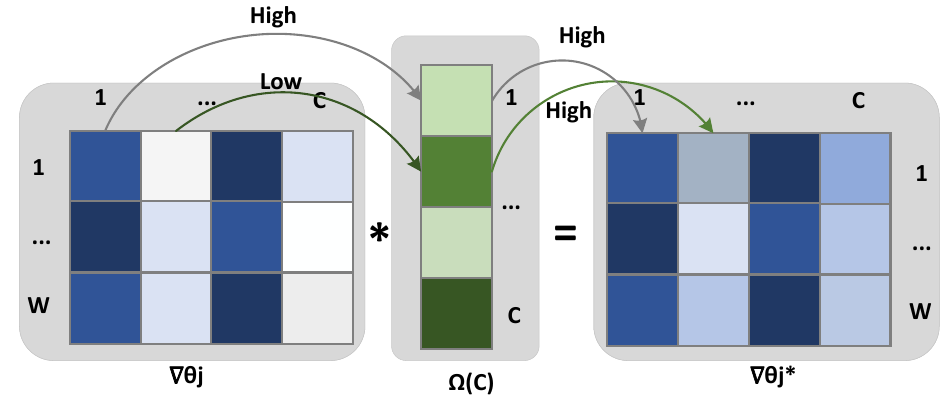}
\vspace{-1em}
  \caption{\small{Details of the function of the gradient weight function. The left side of the image shows that some channels have lower backpropagation gradients, while others have higher ones. Higher backpropagation gradients indicate that these feature channels are being updated more significantly.}}
\vspace{-2.0em}
\label{fig:weight}
\end{figure}

Traditional gradient update strategies adjust model parameters based on global errors, which, although effective, can lead to knowledge interference between different types of tasks. Our new strategy introduces a feature channel weight allocation mechanism, dynamically adjusting the gradient update weights of different feature channels. Specifically, we introduce the weight function $\Omega(k, c)$ implementation:
\vspace{-1em}
\begin{equation}
    \Omega(c_i)  =
    \begin{cases}
        1 + \beta \cdot e^{\frac{iter_{num}}{max_{iter}}}, & \bar{c_i} < \bar{c} ,\\
        1, & \bar{c_i} > \bar{c},
    \end{cases}
\label{oumiga}
\end{equation}
where \(\beta\) is a hyper-parameter, \(iter_{num}\) denotes the current iteration count, and \(max_{iter}\) represents the maximum number of iterations. \(\bar{c}\) is the average across all feature channels, while \(\bar{c_i}\) refers to the average of the \(i\)th feature channel. By dynamically weighting feature channels, we can exert control over the real-time updates of specific feature channels.
This ensures that redundant feature channels can learn newly acquired semantic information without compromising existing reasoning abilities, thereby balancing the enhancement of reasoning abilities and video semantic understanding. As shown in Fig.~\ref{fig:weight}, our gradient weighting function is applied to the backpropagation gradients, boosting those originally lower gradients, thereby substantially increasing the update of channels that were previously updated little or not at all. This process ensures that newly learned semantic information is directed to these newly emphasized feature channels for learning. Therefore, the model parameter update process with the introduction of the RLO optimization strategy is as follows, showing the changes from Eq.~\ref{eq:5} to Eq.~\ref{eq:8}:
\vspace{-0.8em}
\begin{align}
\label{eq:8}
\nabla \theta_j^* &=\Omega(c_i) \cdot \nabla \theta_j , \\
\theta'_j &\leftarrow \theta_j + \nabla \theta_j^*,
\end{align}

During the training process, the input data \( \mathcal{Z} \) consists of input videos and their corresponding questions \( x_i \), correct answers \( y_i \), where \( i = 1, 2, 3, \ldots, K \), and each \( x_i \), \( y_i \) is a sequence of tokens. Large language models are denoted by \( P_\phi(y \mid x) \), where \( \phi \) represents model parameters. Typically, when fine-tuning large models, only the question-answer pairs \( (x_i, y_i) \) are input. The goal is to adjust the parameters \( \phi \) to maximize the following expression:
\vspace{-0.6em}
\begin{equation}
     \max_{\phi} \sum_{(x,y)\in \mathcal{Z}} \sum_{t=1}^{|y|} \log (P_{\phi}(y_t | x, y_{<t})),
\end{equation}
where $|y|$ represents the length of the input vector $y$, $P_{\phi}(y_t | x, y_{<t})$ indicates the likelihood, under model parameters \(\phi\), of predicting the current step \(t\) output \(y_t\) given the input \(x\) and all previously predicted outputs \(y_{<t}\) (i.e., the first \(t-1\) elements of sequence \(y\)). The overall framework is shown in Fig.~\ref{fig:framework}.
\vspace{-0.6em}
\section{Experiments}
\label{experiments}

\begin{table*}[t!]
 \centering
 \resizebox{\textwidth}{!}{
\begin{tabular}{ll|ccccc|cc|ccccc|cc}
  \toprule
  \multirow{3}{*}{Method} & \multirow{3}{*}{LLM}  & \multicolumn{7}{c|}{\bf Zero-shot} &  \multicolumn{7}{c}{\bf Fine-tune} \\
  \cmidrule(lr){3-16}
  ~& ~&\multicolumn{5}{c|}{\bf Countix-QA} & \multicolumn{2}{c|}{\bf  rPPG-QA}  & \multicolumn{5}{c|}{\bf  Countix-QA} & \multicolumn{2}{c}{\bf rPPG-QA} \\ 
  \cmidrule(lr){3-7} \cmidrule(lr){8-9} \cmidrule(lr){10-14} \cmidrule(lr){15-16} 
  & & MAE & RMSE & Bleu1 & CIDEr & METEOR  & MAE & RMSE  & MAE & RMSE & Bleu1 & CIDEr & METEOR  & MAE & RMSE \\
  \midrule
  FrozenBiLM~\cite{frozenbilm} & DeBERTa-V2  & 7.22 & 11.53 &0.145 & 0.550 & 0.112 & 20.43 & 21.56 & 5.22 & 9.44 &0.233 & 0.610 & 0.201 & 19.21 & 20.46\\
  VideoLLaMA~\cite{videollama} & Vicuna-7B  & 6.18 & 10.82 & 0.187 & 0.360 & 0.245 & 18.89 & 19.43 & 4.98 & 8.89 & 0.389 & 0.570 & 0.358 & 18.29 & 19.01\\
  LLaMA-Adapter~\cite{llamaadapter} & LLaMA-7B & 7.44 & 10.51 & 0.244& 0.683 & 0.312 & 21.78 & 23.34 & 5.78 & 10.02 & 0.378& 0.789 & 0.397 & 20.54 & 22.84\\
  VideoChat~\cite{videochat} & Vicuna-7B  & 6.73 & 11.23 &0.165 & 0.549 & 0.277 & 19.54 & 20.67 & 5.03 & 9.53 &0.275 & 0.679 & 0.388 & 18.32 & 10.77\\
  Video-ChatGPT~\cite{videochatgpt} & Vicuna-7B & 6.44 & 10.75 &0.203 & 0.589 & 0.297 & 18.12 & 20.45 & 4.64 & 9.15 &0.356 & 0.643 & 0.367 & 17.54 & 19.45\\
  BT-Adapter~\cite{btadapter} & Vicuna-7B & 7.73 &  11.70 & 0.276 & 0.674 & 0.341 & 19.37 & 22.68 & 6.12 &  9.70 & 0.296 & 0.689 & 0.353 & 18.67 & 21.49 \\
   LLaMA-VID~\cite{llamavid} & Vicuna-7B & 6.97 & 11.21 & 0.267 & 0.680 & 0.301 & 18.31 & 18.56 & 5.34 & 9.85 & 0.388 & 0.783 & 0.411 & 17.51 & 17.60\\
  \midrule
  \rowcolor{mygray}
  {\bf Period-LLM} & LLaMA-7B & {\bf 5.51 } & {\bf 8.33} & {\bf 0.288} & {\bf 0.702} & {\bf 0.308} & {\bf 15.08} & {\bf 17.13} & {\bf 3.77} & {\bf 7.14} & {\bf 0.397} & {\bf 0.810} & {\bf 0.415} & {\bf 13.78} & {\bf 16.83}\\
  \bottomrule
\end{tabular}
 }
 \vspace{-1em}
 \caption{Comparison of our method with other leading methods on two newly introduced video question-and-answer (QA) datasets, with all methods evaluated at a resolution of $224\times 224$. The results for Period-LLM in the zero-shot setting are derived from a model pretrained on text-QA dataset. For rPPG-QA, the V4V dataset is used as the validation set.
 }
 \vspace{-0.3em}
 \label{tab:main_video}
\end{table*}

\begin{table*}[t!]
 \centering
 \resizebox{0.9\textwidth}{!}{
\begin{tabular}{ll|cc|cc|cc|cc|cc|cc}
  \toprule
  \multirow{3}{*}{Method} & \multirow{3}{*}{LLM} & \multicolumn{6}{c|}{\bf Zero-shot} &  \multicolumn{6}{c}{\bf Fine-tune}\\
  \cmidrule(lr){3-14}
  ~& ~&\multicolumn{2}{c|}{\bf RotNIST-QA} & \multicolumn{2}{c|}{\bf Drive-QA} & \multicolumn{2}{c|}{\bf Radar-QA} & \multicolumn{2}{c|}{\bf RotNIST-QA} & \multicolumn{2}{c|}{\bf Drive-QA} & \multicolumn{2}{c}{\bf Radar-QA}\\ 
  \cmidrule(lr){3-4} \cmidrule(lr){5-6} \cmidrule(lr){7-8} \cmidrule(lr){9-10} \cmidrule(lr){11-12} \cmidrule(lr){13-14} 
  ~& ~& MAE & RMSE & MAE & RMSE  & MAE & RMSE & MAE & RMSE & MAE & RMSE  & MAE & RMSE \\
  \midrule
  FrozenBiLM~\cite{frozenbilm} & DeBERTa-V2 & 4.26 & 4.78 &30.88 & 120.56 & 22.17 & 30.42 & 2.88 & 3.25 &29.44 & 110.56 & 16.66 & 28.55\\
  VideoLLaMA~\cite{videollama} & Vicuna-7B  & 4.08 & 5.32 & 35.45 & 130.67 & 23.42 & 18.85 & 2.51 & 3.89 & 31.22 & 121.55 & 17.03 & 20.34 \\
  LLaMA-Adapter~\cite{llamaadapter} & LLaMA-7B  & 3.95 & 4.42 & 31.26 & 125.42 & 24.12 & 31.79 & 2.66 & 3.23 & 30.28 & 118.79 & 20.34 & 28.56 \\
  VideoChat~\cite{videochat} & Vicuna-7B  & 3.56 & 4.21 &32.67 & 126.88 & 21.71 & 29.58 & 2.23 & 3.06 &30.12 & 115.23 & 20.54 & 27.38 \\
  Video-ChatGPT~\cite{videochatgpt} & Vicuna-7B  & 3.44 & 4.87 &38.18 & 118.29 & 23.77 & 32.13 & 2.01 & 2.29 &33.28 & 112.35 & 21.61 & 30.45\\
  BT-Adapter~\cite{btadapter} & Vicuna-7B  & 4.03 &  5.46 & 34.21 & 110.24  & 23.42 & 31.47 & 2.78 &  3.56 & 31.31 & 107.65 & 22.12 & 30.19 \\
   LLaMA-VID~\cite{llamavid} & Vicuna-7B & 3.87 & 4.57 & 34.98 & 114.33 & 18.21 & 29.32 & 2.43 & 3.25 & 32.45 & 109.87 & 18.21 & 29.32 \\
  \midrule
  \rowcolor{mygray}
  {\bf Period-LLM} & LLaMA-7B  & {\bf 3.20} & {\bf 3.96} & {\bf 30.88} & {\bf 108.74}  & {\bf 17.56} & {\bf 26.45}& {\bf 1.50} & {\bf 2.75} & {\bf 28.71} & {\bf 105.54}  & {\bf 14.24} & {\bf 24.12} \\
  \bottomrule
\end{tabular}
 }
 \vspace{-1.0em}
 \caption{Comparison of our method with other leading methods on the RotNIST-QA, Drive-QA, and Radar-QA datasets. Drive-QA is a dataset of taxi trajectory data, while Radar-QA is a radar modality dataset capturing physiological signals. The zero-shot results for Period-LLM are based on a model pretrained on text-QA, Countix-QA, and rPPG-QA. 
 }
 \vspace{-1.7em}
 \label{tab:other_3_exp}
\end{table*}

\begin{table*}[t]
\tiny
\centering
\caption{The MAE value on Countix-QA with different parameter or setting choices in $\Omega(c_{i})$ of RLO optimization strategy, and ablation study on Countix-QA showing the impact of easy tasks (period-text-QA) and RLO (Resisting Logical Oblivion). Default settings are marked in \colorbox{baselinecolor}{gray}.}
\label{tab:combined}
\vspace{-2em}
\subfloat[
\label{tab:beta}
Impact of $\beta$.
]{
\centering
\begin{minipage}{0.15\linewidth} 
\begin{center}
\begin{tabular}{x{10}|x{13}}
\toprule[1pt]
$\beta$        &   MAE            \\
\midrule
0.01    &   3.85           \\
\rowcolor{gray!30}
0.05        &   \textbf{3.77}            \\
0.1        &   3.90           \\
0.5        &   4.05           \\
\bottomrule[1pt]
\end{tabular}
\end{center}
\end{minipage}
}
\hspace{0.3em}
\subfloat[
Type of Threshold.
\label{tab:threshold}
]{
\centering
\begin{minipage}{0.15\linewidth} 
\begin{center}
\begin{tabular}{x{20}|x{15}}
\toprule[1pt]
Threshold   &   MAE            \\
\midrule
\rowcolor{gray!30}
Mean       &   \textbf{3.77}   \\
Median           &   3.85            \\
Learnable              &   4.20           \\
\bottomrule[1pt]
\multicolumn{2}{c}{~}               \\
\end{tabular}
\end{center}
\end{minipage}
}
\hspace{0.3em}
\subfloat[
Results on Countix-QA w./w.o Easy Task and RLO.
\label{tab:abla1}
]{
\centering
\begin{minipage}{0.30\linewidth} 
\begin{center}
\begin{tabular}{x{25}x{28}|x{10}|x{10}x{13}}
\toprule[1pt]
\multicolumn{2}{c|}{\textbf{Easy to Hard}} & \multirow{2}{*}{\textbf{RLO}} & \multicolumn{2}{c}{\textbf{Countix-QA}}  \\
\cmidrule(lr){1-2} \cmidrule(lr){4-5}
\textbf{Text-QA} & \textbf{Countix-QA}  & ~ & \textbf{MAE} & \textbf{CIDEr} \\
\midrule
\xmark & \cmark  & \xmark & 4.30 & 0.661   \\
\cmark & \cmark  & \xmark & 3.89 & 0.782   \\
\rowcolor{gray!30}
\cmark & \cmark  & \cmark & 3.77 & 0.810  \\
\bottomrule[1pt]
\end{tabular}
\end{center}
\end{minipage}
}
\hspace{0.3em}
\subfloat[
Results on rPPG-QA w./w.o Easy Tasks and RLO.
\label{tab:abla2}
]{
\centering
\begin{minipage}{0.36\linewidth} 
\begin{center}
\begin{tabular}{x{25}x{28}x{23}|x{10}|x{23}}
\toprule[1pt]
\multicolumn{3}{c|}{\textbf{Easy to Hard}} & \multirow{2}{*}{\textbf{RLO}} & \multicolumn{1}{c}{\textbf{rPPG-QA}}  \\
\cmidrule(lr){1-3} \cmidrule(lr){5-5}
\textbf{Text-QA} & \textbf{Countix-QA} & \textbf{rPPG-QA} & ~ & \textbf{MAE}  \\
\midrule
\xmark & \xmark & \cmark & \xmark  &  15.45 \\
\cmark & \xmark & \cmark & \xmark &  14.46  \\
\cmark & \cmark & \cmark & \xmark & 14.05  \\
\rowcolor{gray!30}
\cmark & \cmark & \cmark & \cmark & 13.78 \\
\bottomrule[1pt]
\end{tabular}
\end{center}
\end{minipage}
}
\vspace{-1em}
\end{table*}

\subsection{Datasets} 
\textbf{Countix-QA}~\cite{dwibedi2020counting}: A video repetition counting dataset derived from Kinetics, containing 8,757 videos for training, validation, and testing. \textbf{rPPG-QA}: A collection of five rPPG face video datasets (VIPL-HR~\cite{niu2019rhythmnet}, PURE~\cite{stricker2014non}, UBFC-rPPG~\cite{bobbia2019unsupervised}, V4V~\cite{revanur2021first}, and BUAA-MIHR~\cite{xi2020image}), mostly with subjects remaining still, and some with head movements.
\textbf{RotNIST-QA}: A dataset based on MNIST~\cite{lecun1998gradient} for studying rotational invariance, featuring handwritten digit images rotated at various angles. \textbf{Drive-QA}: T\_DRIVE20150206~\cite{libcitylong} GPS trajectory data of taxis in Beijing on February 6, 2015, used for trajectory prediction and urban mobility research. \textbf{Radar-QA}~\cite{schellenberger2020dataset} utilizes the characteristics of vital signs recorded by radar. We verify the algorithm to the radar mode.

\subsection{Model and Metrics} 

\textbf{Training Details.} The model was trained on an NVIDIA A6000 GPU with the Adam optimizer, using an initial learning rate of 0.001 for stable convergence. A batch size of 1 was selected, with the images resized to \(224 \times 224\) pixels. Each video input contained 20 frames, and training spanned 200,000 iterations. Following LLaVA~\cite{liu2024visual_llava}, we used the pretrained CLIP~\cite{radford2021learning} visual encoder ViT-L/14 to extract visual features \( Z_v = g(X_v) \) and applied a linear layer to project image features into the word embedding space via a trainable matrix, transforming \( Z_v \) into language embedding tokens \( H_v \). For Resisting Logical Oblivion, the parameter \( \beta \) was set to 0.05.

\textbf{Metrics.} In addition to content quality, we evaluated the model’s accuracy in predicting cycles of periodic tasks (e.g., action counts, heart rate) using Mean Absolute Error (MAE) and Standard Deviation (STD). For numeric predictions, Arabic and English numerals are identified as \( \text{Num}_{\text{pre}} \), with MAE and STD calculated between predicted and actual numbers. For text quality in natural language processing, we used \textbf{Bleu1} for word-level similarity, \textbf{CIDEr} for caption content relevance, and \textbf{METEOR} to assess semantic similarity with attention to word variation and order.

\subsection{Quantitative Results}
Table~\ref{tab:main_video} compares the performance of the Period-LLM model with other leading methods (FrozenBiLM~\cite{frozenbilm}, VideoLLaMA~\cite{videollama}, LLaMA-Adapter~\cite{llamaadapter}, VideoChat~\cite{videochat}, Video-ChatGPT~\cite{videochatgpt}, BT-Adapter~\cite{btadapter}, LLaMA-VID~\cite{llamavid}) on two video question-and-answer datasets (Countix-QA and rPPG-QA). 
In the zero-shot protocol, other large multimodal models (MLLMs) make direct predictions, while Period-LLM first undergoes pretraining on a text-QA dataset. Results show that Period-LLM outperforms existing MLLMs, demonstrating that pretraining with periodic text tasks enhances the model's ability to understand periodicity.
For fine-tuning, other MLLMs are directly fine-tuned on the Countix-QA and rPPG-QA datasets using open-source code. In contrast, Period-LLM adopts an easy-to-hard generalization approach, combined with a Resisting-Logical-Oblivion strategy. Fine-tuning is done progressively from easier to harder tasks, which helps retain periodic reasoning abilities. The results show significant improvements over zero-shot, thanks to the inclusion of explicit periodic information in the training data.
Period-LLM surpasses existing MLLMs by over 1 MAE on the Countix-QA dataset, while also showing significant improvements in generated text quality, as measured by Bleu-1, CIDEr, and METEOR scores. This improvement is driven by the easy-to-hard fine-tuning paradigm and the Resisting-Logical-Oblivion strategy, which together help the model preserve and enhance its periodic reasoning abilities, while also maintaining strong semantic understanding, resulting in superior performance.

In addition to the visual periodic tasks, Table~\ref{tab:other_3_exp} presents the performance of these models on other modalities, specifically the RotNIST-QA, Drive-QA, and Radar-QA datasets. The zero-shot and fine-tune protocols follow the same approach as in Table~\ref{tab:main_video}, with the key difference being that experiments on these three datasets were conducted in parallel. The pretraining models used in these experiments were fine-tuned on the two visual periodic tasks from Table~\ref{tab:main_video}.
On the RotNIST-QA dataset, Period-LLM achieves the lowest MAE (1.50) and RMSE (2.75), outperforming models such as FrozenBiLM, VideoLLaMA, and LLaMA-Adapter. For the Drive-QA dataset, Period-LLM demonstrates superior performance with the lowest MAE (28.71) and competitive RMSE (105.54). On the Radar-QA dataset, Period-LLM also performs strongly, further validating the effectiveness of our model across different modalities. These results underscore Period-LLM's enhanced accuracy in periodic reasoning tasks and its robustness in handling diverse datasets, establishing it as a versatile solution for periodic question-answering across a range of domains, including health, weather, and traffic.

\subsection{Qualitative Results} 
\begin{figure}
\centering
\vspace{-2em}
\includegraphics[width=0.8\linewidth]{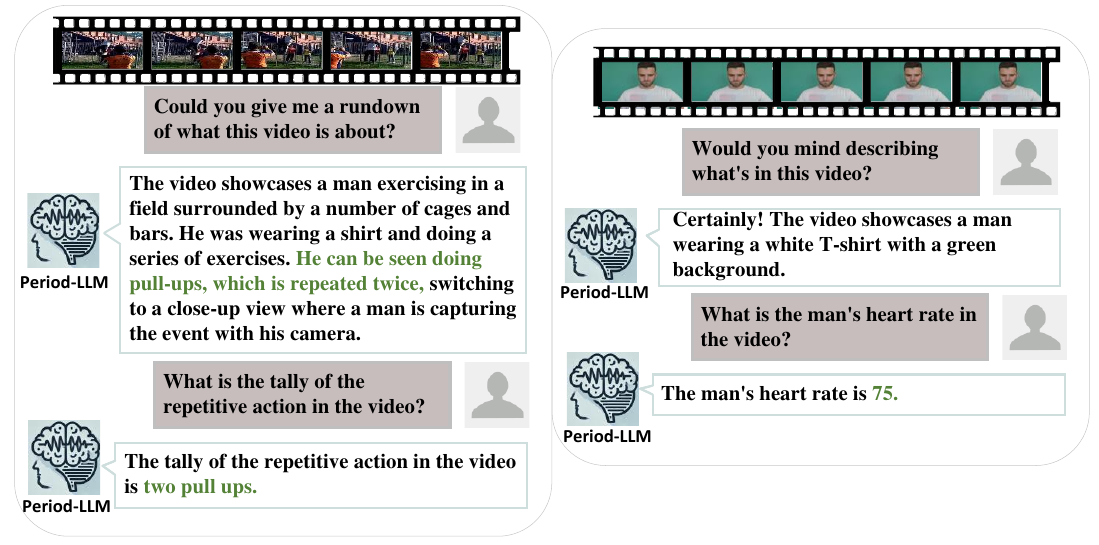}
\vspace{-1em}
  \caption{\small{Examples from Countix-QA and rPPG-QA.
  }
  }
\vspace{-1em}
\label{fig:QA}
\end{figure}

\textbf{QA Examples.} Fig.~\ref{fig:QA} showcases examples from the countix-QA and rPPG-QA datasets. It can be seen that Period-LLM is capable of understanding the semantic information contained within the video as well as recognizing periodicity within the video. For instance, in the countix-QA dataset, Period-LLM accurately provides the number of repetitive movements, while in the rPPG-QA dataset, it can determine the heart rate of the man in the video. These capabilities are beyond what other language models can achieve. According to Fig.~\ref{fig:QA}, in the countix-QA dataset, Period-LLM identifies that the man in the video performs two pull-ups. In the rPPG-QA dataset, Period-LLM is able to detect that the man’s heart rate is 75 beats per minute. This demonstrates Period-LLM's exceptional performance in handling video content and periodic information. More QA examples are provided in the appendix.

\begin{figure}
\centering
\includegraphics[width=0.8\linewidth]{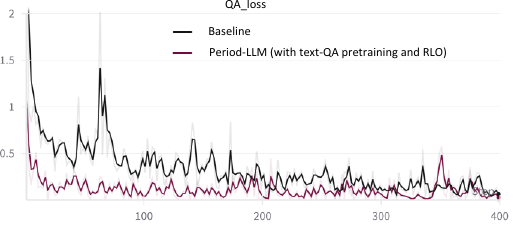}
\vspace{-1em}
  \caption{\small{Comparison of Loss Reduction Between Baseline (Vanilla Fine-Tuning on Countix-QA) and Proposed Method (Period-LLM with Text-QA Pretraining and RLO, Fine-Tuned on Countix-QA).
  }
  }
\vspace{-1.5em}
\label{fig:loss}
\end{figure}

\vspace{-0.2em}
\textbf{Comparison of Loss Reduction.} The Fig.~\ref{fig:loss} illustrates the loss reduction comparison between the Baseline (Vanilla Fine-Tuning on Countix-QA) and the proposed Period-LLM approach (with text-QA pretraining and RLO) on the Countix-QA dataset. As shown, our proposed method demonstrates a significantly faster decline in the loss values compared to the baseline. The Period-LLM model achieves a steeper descent in the initial training stages, indicating a more efficient learning process. Furthermore, it converges to a lower loss value in fewer epochs, suggesting improved optimization and stability. This faster convergence implies that our method better captures the underlying data patterns, leading to more effective fine-tuning on the Countix-QA.

\vspace{-0.8em}
\subsection{Ablation Study}
\textbf{The Setting of Weight Function $\Omega(c_{i})$ of RLO Optimization Strategy.} As seen in Table \ref{tab:beta}, $\beta = 0.05$ yields the lowest MAE (3.77), indicating an effective balance between retaining prior knowledge and integrating new information. When $\beta$ is too small (e.g., 0.01), the model’s anti-forgetting ability is weak, causing higher error as new knowledge overwhelms older knowledge. Conversely, a high $\beta$ (e.g., 0.5) overly prioritizes old knowledge, limiting adaptation. Thus, $\beta = 0.05$ achieves the best balance for minimizing error.
In Table \ref{tab:threshold}, setting the threshold to the mean results in the lowest MAE (3.77), effectively balancing old and new knowledge contributions. The median threshold performs slightly worse (MAE 3.85) due to its insensitivity to outliers, while a learnable threshold increases MAE (4.20), likely due to instability. Hence, the mean threshold is the most stable and effective choice.

\textbf{Results of various combinations of Preiod-LLM Components.} We evaluate our approach on the Countix-QA and rPPG-QA datasets, as shown in Tables~\ref{tab:abla1} and~\ref{tab:abla2}. For Countix-QA, our method—combining pre-training on simpler tasks with the Resisting Logical Oblivion (RLO) strategy—significantly improves performance, reducing the Mean Absolute Error (MAE) from 4.30 to 3.77 and increasing the CIDEr score from 0.661 to 0.810. Similarly, on the rPPG-QA dataset, this approach lowers the MAE from 15.45 to 13.78. These improvements reflect the model’s enhanced capacity to capture periodic patterns effectively while also understanding semantic content more deeply. The decrease in MAE shows better accuracy in numerical predictions, while the CIDEr increase suggests improved semantic alignment in question-answering tasks. Together, these results confirm that the integration of easy-task pre-training with RLO enables the model to balance periodicity and semantic understanding, leading to faster convergence and more robust performance across QA tasks.
\vspace{-1em}
\section{Conclusion}
In this paper, we introduce Period-LLM, a large multimodal model specifically designed to enhance performance on periodic tasks across various modalities. The model follows a progressive training paradigm, starting with simpler text-based tasks and advancing to more complex visual and multimodal tasks. Additionally, we propose the ``Resisting Logical Oblivion” optimization strategy to preserve the model's periodic reasoning abilities during semantic alignment.the 
Our experimental results demonstrate that Period-LLM significantly outperforms existing models in handling periodic tasks, especially in challenging domains such as repetitive motion analysis and physiological signal estimation. Period-LLM not only achieves superior accuracy but also highlights its potential for future advancements in managing complex periodic phenomena across different modalities.
Future work will focus on further refining model's capabilities, particularly in addressing more nuanced and less structured periodic tasks, and expanding its application to additional domains where periodicity plays a crucial role.
\section{Acknowledgment}

This work was supported by the Guangdong Provincial Key Laboratory of Integrated Communication, Sensing and Computation for Ubiquitous Internet of Things (No. 2023B1212010007), China NSFC Grant 62472366, the Project of DEGP (No.2024GCZX003, 2023KCXTD042), 111 Center (No.D25008) and Shenzhen Science and Technology Foundation (ZDSYS20190902092853047).

{
    \small
    \bibliographystyle{ieeenat_fullname}
    \bibliography{main}
}
\end{document}